\documentclass[letterpaper, 10 pt, conference]{ieeeconf}  

\IEEEoverridecommandlockouts                            

\usepackage{times}
\usepackage{epsfig}
\usepackage{graphicx}
\usepackage{amsmath}
\usepackage{amssymb}
\usepackage{array}
\usepackage{multirow}
\usepackage{epsfig}
\usepackage{booktabs}
\usepackage{tabularx}
\usepackage{color}
\usepackage{subfigure}
\usepackage{bm}
\usepackage{diagbox}
\usepackage[colorlinks=true,linkcolor=blue,citecolor=green,urlcolor=blue,]{hyperref}
\usepackage[table,xcdraw]{xcolor}

\definecolor{gray}{RGB}{229, 230, 230}
\usepackage[ruled,linesnumbered]{algorithm2e}

\title{\LARGE \bf
Adaptive Stereo Depth Estimation with Multi-Spectral Images Across All Lighting Conditions
}

\author{Zihan Qin, Jialei Xu, Wenbo Zhao, Junjun Jiang and Xianming Liu$^*$
\thanks{Zihan Qin, Jialei Xu, Wenbo Zhao, Junjun Jiang, and Xianming Liu are with the School of Computer Science and Technology, Harbin Institute of Technology, Harbin, 150001, China, E-mail: \{24s103315, xujialei\}@stu.hit.edu.cn, \{wbzhao, jiangjunjun, csxm\}@hit.edu.cn. $^*$Corresponding author.} } 

\begin{document}

\maketitle
\thispagestyle{empty}
\pagestyle{empty}

\begin{abstract}

Depth estimation under adverse conditions remains a significant challenge. Recently, multi-spectral depth estimation, which integrates both visible light and thermal images, has shown promise in addressing this issue. However, existing algorithms struggle with precise pixel-level feature matching, limiting their ability to fully exploit geometric constraints across different spectra. To address this, we propose a novel framework incorporating stereo depth estimation to enforce accurate geometric constraints. In particular, we treat the visible light and thermal images as a stereo pair and utilize a Cross-modal Feature Matching (CFM) Module to construct a cost volume for pixel-level matching. To mitigate the effects of poor lighting on stereo matching, we introduce Degradation Masking, which leverages robust monocular thermal depth estimation in degraded regions. Our method achieves state-of-the-art (SOTA) performance on the Multi-Spectral Stereo (MS2) dataset, with qualitative evaluations demonstrating high-quality depth maps under varying lighting conditions.

\end{abstract}

\section{INTRODUCTION}

Depth estimation has received considerable attention in recent years due to its widespread applications in areas such as autonomous driving~\cite{wang2019pseudo,you2020pseudo}, robotics~\cite{wofk2019fastdepth}, and 3D reconstruction~\cite{geiger2011stereoscan}. Significant advancements have been made in both monocular and stereo depth estimation through deep learning-based approaches. However, these depth estimation algorithms predominantly rely on the visible domain. Consequently, their performance often suffers significant degradation due to the decline in image quality, particularly under poor illumination conditions such as nighttime or rainy weather~\cite{shin2023deep}, which prevented these algorithms from being widely applied in real-world scenarios.

To address this challenge, recent research has increasingly investigated alternative vision modalities such as near-infrared images~\cite{park2022adaptive,brucker2024cross}, 
and long-wave infrared (also known as thermal) images~\cite{shin2023deep,lu2021alternative,guo2023unsupervised} to achieve reliable and robust depth estimation in adverse conditions. Among these alternative modalities, thermal images have gained more popularity due to their low acquisition cost, robustness in adverse conditions, and consistent performance regardless of lighting variations. Some researchers have attempt to achieve depth estimation with only thermal images~\cite{shin2023deep,lu2021alternative}. However, thermal images typically exhibit lower texture information, resolution, and higher noise levels compared to visible light images, resulting in less accurate depth estimation in well-lit areas.

Given that thermal and visible light depth estimation each have distinct strengths and weaknesses, researchers have focused on integrating thermal images with visible light images to leverage complementary information from both modalities. However, the significant differences between thermal and visible images present challenges in exploiting correlations across modalities. The substantial appearance differences, pacrtiularly the lack of texture in thermal images, complicate keypoint matching. Additionally, low illumination can obscure objects in visible light images, leading to mismatches with thermal images. Consequently, previous multi-spectral methods~\cite{guo2023unsupervised,shin2023self} generally avoid direct pixel-level matching between images, limiting their ability to leverage geometric constraints. As a result, their performance is highly dependent on the training dataset and exhibits poor generalization.


\begin{figure}[t!]
    \centering
    \subfigure{
    \includegraphics[width=0.97\linewidth]{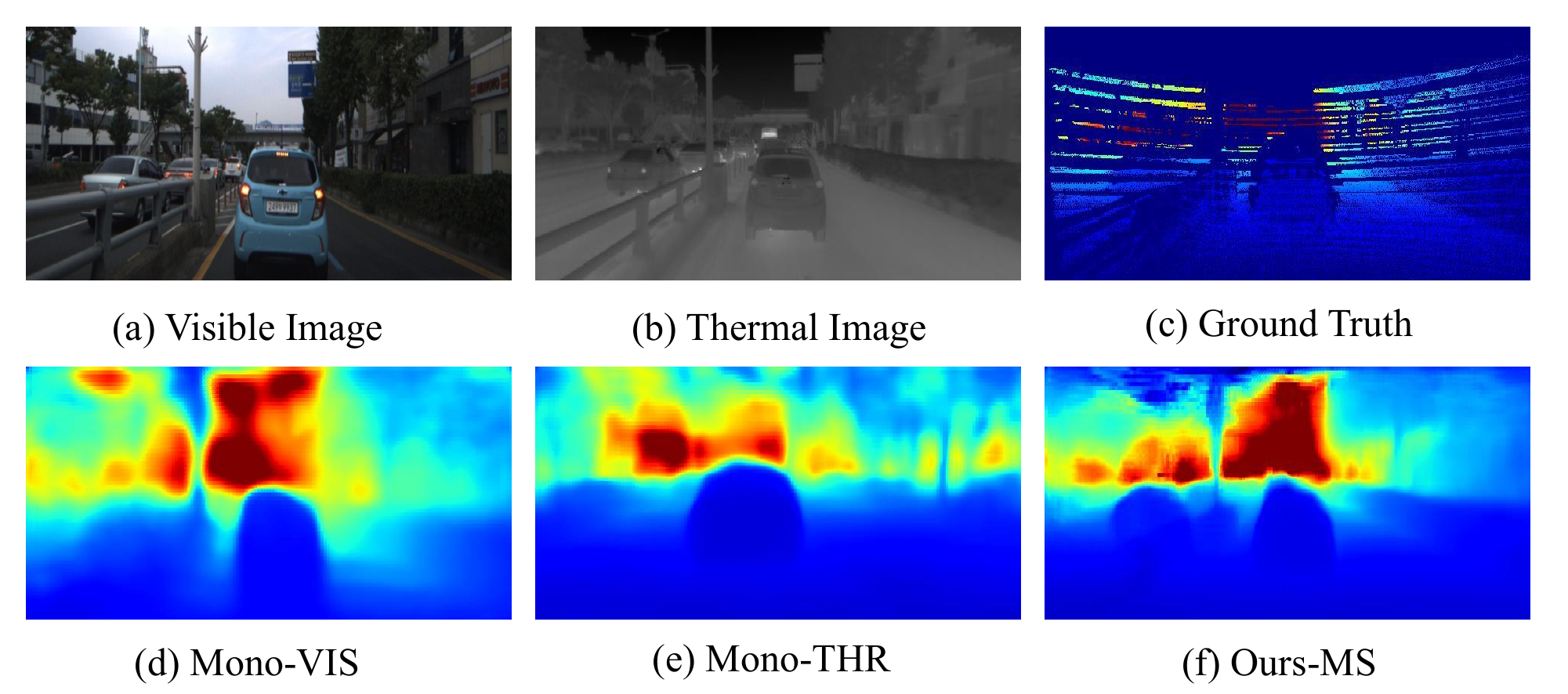}
    }
    \vspace{-10pt}
    \caption{\textbf{Depth from images of different modalities.}  (a) and (b) show the visible light and thermal images, respectively; (c) is the LiDAR ground truth corresponding to the thermal image; (d) and (e) present the depth maps obtained from monocular methods using the visible light and thermal modalities, respectively; (f) illustrates the depth map estimated by our multi-spectral method.}
    \vspace{-10pt}
    \label{fig:fig1}
\end{figure}

To this end, we propose a novel framework that integrates thermal and visible light images for robust and accurate depth estimation under varying lighting conditions, as shown in Fig.~\ref{fig:fig1}. Specifically, we first treat the input visible light and thermal images as a stereo pair and train the Cross-modal Feature Matching (CFM) Module to generate aligned feature vectors for each pixel. This alignment allows us to project visible light features onto thermal features across candidate depths, constructing a cost volume that provides accurate pixel-level matching in well-lit regions. Then we estimate depth probability distributions for both modalities, and introduce a degradation mechanism based on those distributions, which reverts to monocular thermal image depth estimation for regions with adverse conditions. Finally, we employ a Depth Module to generate the final depth map. Our experimental evaluations demonstrate that our method surpasses existing state-of-the-art depth estimation methods, marking a significant advancement in the field. Our contributions are summarized as follows:

\begin{itemize}

\item We propose a novel multi-spectral depth estimation method that leverages pixel-level matching through stereo depth estimation. By incorporating geometric constraints between cameras, our method achieves more accurate depth estimation. To the best of our knowledge, this is the first application of stereo depth estimation architecture in this domain.

\item For low-light regions where the visible light image becomes unreliable, we introduce a novel degradation mechanism that effectively degrades to monocular thermal depth estimation.

\item We demonstrate that the proposed method achieves substantial improvements over current state-of-the-art techniques on the MS2 benchmark dataset~\cite{shin2023deep}.

\end{itemize}


\section{RELATED WORKS}

\subsection{Depth Estimation in the Visible Light Domain}
\textbf{Monocular Depth Estimation} aims to infer scene depth from a single image. It can be approached as either a regression problem or a classification problem. Regression-based methods~\cite{eigen2014depth,laina2016deeper,fu2018deep,xu2022multi} involve predicting per-pixel depth values using convolutional neural networks. In contrast, classification-based approaches~\cite{bhat2021adabins,bhat2022localbins,shao2024iebins} divide depth into discrete intervals and predict probabilities for each pixel, transforming the depth estimation task into a classification problem.

However, monocular depth estimation is an ill-posed problem because a single 2D image can correspond to an infinite number of different 3D scenes. As a result, estimating absolute depth often leads to overfitting on specific datasets, capturing patterns that may not generalize to new environments~\cite{bhat2023zoedepth}. Meanwhile, estimating only relative depth provides limited practical value in real-world applications.

\textbf{Stereo Depth Estimation} involves estimating a pixel-wise disparity map from a stereo image pair, which can be used to determine the depth of each pixel in the scene. Learning-based stereo depth estimation methods can roughly be divided into two main categories: the encoder-decoder network with 2D convolution~\cite{mayer2016large,ilg2018occlusions,weinzaepfel2023croco,xu2024sdge} and the cost volume matching with 3D convolution~\cite{chang2018pyramid,zhang2019ga,zhang2020domain,xu2023iterative}. The former directly outputs a disparity map, while the latter requires feature matching at multiple disparity levels to construct a cost volume, leading to higher computational costs but typically achieving greater accuracy.

Compared to monocular depth estimation, stereo depth estimation benefits from the geometric constraints between two viewpoints, resulting in significantly improved accuracy. However, stereo depth estimation faces challenges in handling occlusions, textureless areas, and reflective surfaces.

\textbf{Multi-view Stereo (MVS)} estimates the dense depth map from overlapping images. Yao et al.~\cite{yao2018mvsnet} extract deep visual features from the input images, followed by the construction of a 3D cost volume via differentiable homography warping. Subsequent works have largely adopted these steps. Yao et al.~\cite{yao2019recurrent} replaces the 3D CNN used in cost volume regularization with GRUs to reduce memory consumption, and similarly, Xu et al.~\cite{xu2021non} utilize a non-local RNN to achieve this objective. Yang et al.~\cite{yang2020cost} have explored the use of feature pyramid networks to extract multi-scale features, enabling a coarse-to-fine construction of the cost volume. Bae et al.~\cite{bae2022multi} integrate monocular depth estimation with Multi-view Stereo techniques, using a depth consistency constraint to ensure alignment between the cost volume and monocular depth, effectively addressing occlusion and textureless surface challenges.

Unlike stereo matching methods, MVS methods do not require stereo rectification of input images, making its approach to constructing the cost volume more generalizable.

\subsection{Multi-spectral Depth Estimation}

Due to the decreased accuracy of depth estimation methods based on visible light under poor lighting conditions, researchers have increasingly turned to multi-spectral images for depth estimation. Treible et al.~\cite{treible2017cats} attempted pixel-level matching between thermal and visible light images but were unable to obtain effective disparity and depth maps because of significant distribution differences between the modalities. 
Shin et al.~\cite{shin2023self} employed an Adversarial Multi-spectral Adaptation method, using visible light images as auxiliary supervision to estimate the depth of thermal images. Kim et al.~\cite{kim2018multispectral} and Lu et al.~\cite{lu2021alternative} utilized specialized hardware to create accurately aligned visible-thermal image pairs for training, yet they still faced limitations in achieving accurate depth estimation due to a lack of geometric constraints. Guo et al.~\cite{guo2023unsupervised} employed cross-spectrum spatial consistency between visible light and thermal images for self-supervised learning. Lastly, Shin et al.~\cite{shin2023deep} proposed a conditional random field block to estimate depth from thermal image pairs. However, due to the inherent limitations of thermal images, their method performs worse than visible light methods under favorable lighting conditions.

Compared to existing multi-spectral depth estimation methods, our approach effectively capitalizes on the complementary strengths of thermal and visible light images. In areas with favorable illumination, we employ geometric constraints derived from stereo vision to achieve precise depth estimation. In contrast, for regions with insufficient illumination, our method transitions to a monocular depth estimation approach based on thermal images. This adaptive mechanism ensures robust and accurate depth estimation across varying lighting conditions, addressing the limitations inherent in prior methodologies.

\begin{figure*}[ht]
\vspace{5pt}
    \centering
    \subfigure{
    \includegraphics[width=0.97\linewidth]{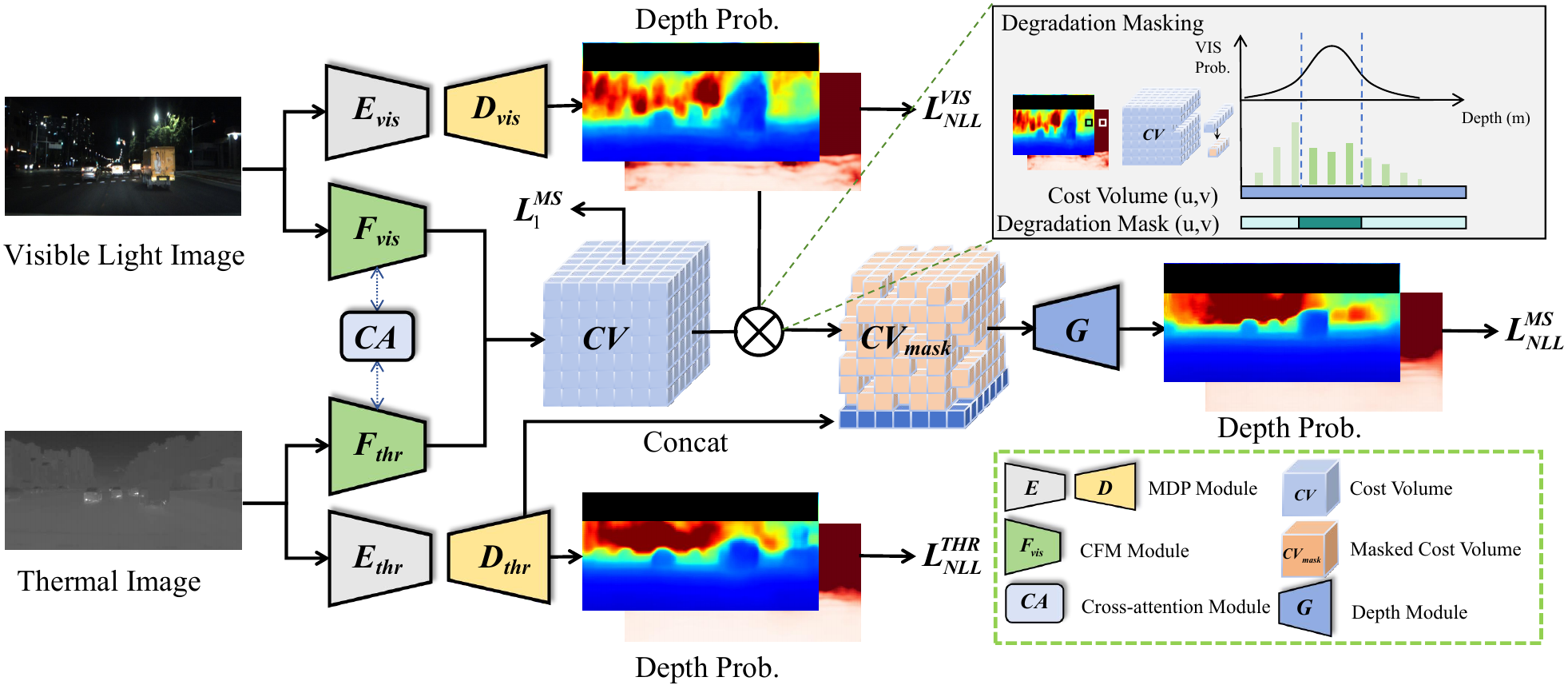}
    }
    \caption{\textbf{The architecture of our proposed network.} The method involves the following steps: First, based on the input visible light images $I_{vis}$ and thermal images $I_{thr}$, we utilize a cross-attention-based feature extractor to generate aligned feature vectors for each pixel by projecting between the two modalities and obtaining the cost volume for pixel-level matching. Second, we perform monocular depth estimation independently for each modality, producing pixel-wise depth probability distributions. Third, we apply Degradation Masking, derived from the visible image's depth probability distribution, to the cost volume to remove inaccurate matches. Finally, we utilize the final layer of features from the thermal MDP Module to degrade the masked cost volume into monocular thermal depth estimation, producing the final depth map through the Depth Module.}
    \label{fig:network}
    \vspace{-10pt}
\end{figure*}

\section{Method}
\subsection{Problem Formulation}
\label{subsec:formulation}
Given a pair of calibrated visible light and thermal images, $\mathbf{I}_{\text{vis}}$ and $\mathbf{I}_{\text{thr}}$, with their respective intrinsics, rotations and translations $\{\mathbf{K}_{\text{vis}}, \mathbf{R}_{\text{vis}}, \mathbf{t}_{\text{vis}} \}$ and $\{\mathbf{K}_{\text{thr}}, \mathbf{R}_{\text{thr}}, \mathbf{t}_{\text{thr}} \}$, our goal is to find a accurate depth estimation from the two images by leveraging their complementary information. The problem can be formulated as finding a function $f(\cdot|\boldsymbol{\theta})$ that maps the input images and camera parameters to the depth map:
\begin{equation}
    \mathbf{D}_{\text{map}} = f(\mathbf{I}_{\text{vis}}, \mathbf{I}_{\text{thr}}, \mathbf{K}_{\text{vis}}, \mathbf{K}_{\text{thr}}, [\mathbf{R}_{\text{vis}}, \mathbf{t}_{\text{vis}}], [\mathbf{R}_{\text{thr}}, \mathbf{t}_{\text{thr}}]|\boldsymbol{\theta}),
\end{equation}
which is supervised by ground truth depth maps obtained through LiDAR measurements. As illustrated in Fig.~\ref{fig:network}, our proposed approach encompasses three principal phases: 

\begin{itemize}

    \item \textbf{Cross-modal Feature Matching.} We first treat the input $\mathbf{I}_{vis}$ and $\mathbf{I}_{thr}$ as a stereo pair and train two Cross-modal Feature Matching (CFM) Modules to generate aligned feature vectors for each pixel. The visible light features are projected onto the thermal features across candidate depths, constructing a cost volume that facilitates pixel-level matching.

    \item \textbf{Degradation Masking.} To address the challenges posed by low-light regions that are difficult to match, we employ a Modality-specific Depth Probability (MDP) Module to estimate the depth probability of each pixel in $\mathbf{I}_{vis}$ as Gaussian distribution, and generate a degradation mask based on the probability. Then the mask is applied on the cost volume, which allows us to remove inaccurate matches.



    
    
    \item \textbf{Depth Map Generation.} Finally, to compensate for the removed portions, we employ another MDP Module to extract features from the thermal image and concatenate them with the masked cost volume. This is then fed into the Depth Module to produce the final depth estimation, allowing regions where matching fails to degrade to monocular thermal depth estimation, providing robust results under varying lighting conditions.


\end{itemize}


\subsection{Cross-modal Feature Matching}



We leverage Multi-view Stereo (MVS) methods to construct a cost volume for feature matching across different views. Initially, we generate feature vectors for each pixel in both views using the CFM Module based on the PSMNet backbone~\cite{chang2018pyramid}. Considering the different modalities of input images, their features do not reside in the same feature space, which hinders subsequent feature matching. To address this, we introduce a Cross-Attention Module to align the feature spaces. This module consists of two cross-attention layers. In the first layer, visible light features are used as queries and thermal features as keys, while in the second layer, the roles are reversed, with thermal features as queries and visible light features as keys:
\begin{equation}
    \mathbf{f}_{aligned} = \text{softmax}\left(\frac{\mathbf{Q} \mathbf{K}^\top}{\sqrt{d}}\right) \mathbf{f}_{origin},
\end{equation}
where $\mathbf{Q}$ and $\mathbf{K}$ are the query vector and key vector, respectively, \(d\) is the dimensionality of the key vector.

These aligned feature vectors are then projected across views by utilizing the intrinsic $\mathbf{K}_{\text{thr}}$, $\mathbf{K}_{\text{vis}}$ and extrinsic parameters $[\mathbf{R}_{\text{vis}},\mathbf{t}_{\text{vis}}]$, $[\mathbf{R}_{\text{thr}},\mathbf{t}_{\text{thr}}]$ to compute the matching pixels. The similarity between features of corresponding pixels is subsequently calculated to construct the cost volume. Specifically, for each pixel \((u, v)\) in $\mathbf{I}_{thr}$, we find its corresponding pixel \((u^{\prime}, v^{\prime})\) in $\mathbf{I}_{vis}$ and select a set of uniformly sampled depth candidates \(\{d_k\}_{k=1}^N\). For each depth candidate \(d_k\), the matching score for depth candidate \(d_k\) is calculated by taking the dot product of the features from both views:

\begin{equation}
C(u, v, d_k) = \mathbf{f}_{thr}(u, v) \cdot \mathbf{f}_{vis}(u^{\prime}, v^{\prime}, d_k),
\end{equation}
where \(\mathbf{f}_{thr}\) and \(\mathbf{f}_{vis}\) are the feature vectors extracted from the thermal and visible light views, respectively. These computed similarities are organized into a cost volume, which is then processed with a softmax operation to generate a depth probability volume:
\begin{equation}
    D(u, v) = \{ P(u, v, d_1), P(u, v, d_2), \dots, P(u, v, d_{N}) \},
\end{equation}
where \( P(u, v, d_k) \) represents the probability of depth \( d_k \) at pixel \((u, v)\), and \( N \) is the number of depth candidates.

\subsection{Degradation Masking}
The Multi-view Stereo matching can provide accurate matching in well-lit regions. However, extracting visible light information in regions with adverse conditions is challenging, which can lead to unreliable matches. To address this problem, we propose a novel strategy, namely degradation masking, to remove inaccurate matches from the cost volume, and degrade them to monocular thermal depth estimation.

Specifically, we firstly identify the regions of low reliability within the visible light modality. To achieve this goal, we compute the depth probability distribution of $\mathbf{I}_{vis}$, as the probability can be a good representation of reliability. Here we employ the D-Net in MaGNet~\cite{bae2022multi} as the MDP model to predict the depth value \(d_{uv}\) for each pixel \((u, v)\) in $\mathbf{I}_{vis}$, and model their probability ($P$) as a Gaussian distribution to capture the depth uncertainty:

\begin{equation}
P(d_{uv}) = \frac{1}{\sqrt{2\pi\sigma_{uv}^2}} \exp\left(-\frac{(d_{uv} - \mu_{uv})^2}{2\sigma_{uv}^2}\right),
\end{equation} 
where \( \mu_{uv} \) represents the mean depth, and \( \sigma_{uv}^2 \) denotes the variance at pixel \((u, v)\).

Since the low probability indicates that the corresponding pixel is more likely to be mismatch. We can simply remove the depth candidate corresponding to low probability to achieve degradation masking. This is achieved by exclude the depth candidate \(d_k\) for that pixel if \(P_{vis}(d_k | (u, v))\) is below a certain threshold $\theta_{(u, v)}$, which is computed by:
\begin{equation}
\theta_{(u, v)}=\mu_{uv}+k*\sigma_{uv},
\end{equation}
where $k$ is a hyperparameter. We have found that setting $k=1$ yields satisfactory results.

\subsection{Depth Map Generation}
To degrade the mismatch of poorly lit regions to monocular thermal depth estimation, we utilize an additional MDP model to predict the depth of $\mathbf{I}_{thr}$. Features extracted from the final layer of this MDP module are concatenated with the masked cost volume. As discussed in~\cite{bae2022multi,chang2018pyramid}, maintaining the size of depth estimation and cost volume at $(H/4, W/4)$ ensures both computational efficiency and accuracy. To generate the depth map and recover the final depth map at full resolution, we apply our proposed Depth Module, which incorporates the learnable upsampling method introduced by Bae et al.~\cite{bae2022multi}:
\begin{equation}
[\mu_d, \sigma_d^2] = \text{DepthModule}(\text{Concat}(C, F_{\text{thermal}})),
\end{equation}
where \( C \) denotes the cost volume constructed by the CFM Module, and \( F_{\text{thermal}} \) represents the feature map from the last layer of the thermal MDP Module. Depth Module outputs the depth mean \(\mu_d\) as the final estimated depth, while the variance \(\sigma_d^2\) is used only during training for loss computation.




\subsection{Training Details}

We divide the training process of the whole network into three stages: CFM Module training, MDP Module training, and Depth Module training. The order of training the CFM and MDP Modules can be interchanged.

In the training of the CFM Module, we multiply the cost volume output by the depth candidates to obtain the expected depth values. The L1 loss is then computed between these expected depth values and the ground truth:
\begin{equation}
    L^{MS}_1=\sum^W_u\sum^H_v\left|\sum^N_{k=1}d_k\cdot p(d)-d^{gt}_{uv}\right|,
\end{equation}
where $d_{uv}$ represents the depth value at pixel ($u,v$), $H$ and $W$ represent the height and width of the image, respectively.

In the training of the MDP Module,  we use the encoder pre-trained on the KITTI dataset~\cite{Geiger2013IJRR} from AdaBins~\cite{bhat2021adabins}, and employ the Negative Log-Likelihood (NLL) loss to optimize the mean and variance for each modality separately.
\begin{equation}
\label{eq:nll}
L_{NLL} = \sum_{u}^W\sum_{v}^H \left[ \frac{(d_{uv}^{gt} - \mu_{uv}(\mathbf I_t))^2}{2 \sigma_{uv}^2(\mathbf I_t)} + \frac{1}{2}\log  \sigma_{uv}^2(\mathbf I_t) \right],
\end{equation}
where $\mu_{uv}$ denotes the predicted mean depth, and $\sigma_{uv}^2$ indicates the variance. Since there are two MDP Modules, we compute two loss $L^{VIS}_{NLL}$, $L^{THR}_{NLL}$ for $\mathbf{I}_{\text{vis}}$ and $\mathbf{I}_{\text{thr}}$, respectively.

In the training of Depth Module, the weights of the MDP Module and CFM Module are frozen. The training is conducted using the same NLL loss function in Eq. \ref{eq:nll} as $L^{MS}_{NLL}$ . The final depth map $D_{map}$ is obtained as the mean $\mu$ derived from this process.
\section{EXPERIMENTS}
\begin{table*}[htbp]
\vspace{5pt}
  \centering
  \caption{Quantitative depth comparison on the official split of MS2 dataset}
  \resizebox{0.95\textwidth}{!}{
  \begin{tabular}{|c|c|cccc|ccc|}
    \hline
    \multirow{2}{*}{\textbf{Method}} & \multirow{2}{*}{\textbf{TestSet}} & \multicolumn{4}{c|}{\textbf{Error} \(\downarrow\)} & \multicolumn{3}{c|}{\textbf{Accuracy} \(\uparrow\)} \\
    \cline{3-9}
    &  & Abs Rel & Sq Rel & RMSE & RMSE log & \(\delta < 1.25\) & \(\delta < 1.25^2\) & \(\delta < 1.25^3\) \\
    \hline
     \multicolumn{1}{|c|}{\multirow{4}{*}{DORN~\cite{fu2018deep}}}
          & day   & 0.144 & 1.288 & 5.483 & 0.230  & 0.856 & 0.941 & 0.970 \\
          & night & 0.136 & 1.136 & 5.290  & 0.212 & 0.863 & 0.950  & 0.976 \\
          & rain  & 0.180  & 1.934 & 6.735 & 0.276 & 0.781 & 0.910  & 0.955 \\
          & \cellcolor{gray}avg & \cellcolor{gray}0.151 &\cellcolor{gray}1.419 & \cellcolor{gray}5.776 & \cellcolor{gray}0.237 &\cellcolor{gray}0.837 &\cellcolor{gray}0.935 &\cellcolor{gray}0.968 \\
    \hline
    \multicolumn{1}{|c|}{\multirow{4}{*}{BTS~\cite{lee2019big}}} & day   &  0.122 & 0.905 & 4.923 & 0.198 & 0.857 & 0.951 & 0.980 \\
          & night & 0.114 & 0.798 & 4.701 & 0.184 & 0.870  & 0.959 & 0.984 \\
          & rain  & 0.157 & 1.395 & 6.053 & 0.243 & 0.791 & 0.926 & 0.969 \\
          &\cellcolor{gray}avg &\cellcolor{gray}0.129 &\cellcolor{gray}1.008 &\cellcolor{gray}5.169 &\cellcolor{gray}0.206 &\cellcolor{gray}0.843 &\cellcolor{gray}0.947 &\cellcolor{gray}0.978 \\
    \hline
    \multicolumn{1}{|c|}{\multirow{4}{*}{AdaBins~\cite{bhat2021adabins}}} & day   & 0.129 & 0.976 & 5.108 & 0.205 & 0.847 & 0.947 & 0.979 \\
          & night & 0.119 & 0.822 & 4.749 & 0.187 & 0.864 & 0.958 & 0.984 \\
          & rain  & 0.168 & 1.545 & 6.336 & 0.254 & 0.771 & 0.918 & 0.965 \\
          &\cellcolor{gray}avg &\cellcolor{gray}0.137 & \cellcolor{gray}1.084 &\cellcolor{gray}5.330  &\cellcolor{gray}0.212 &\cellcolor{gray}0.831 & \cellcolor{gray}0.943 &\cellcolor{gray}0.977 \\
    \hline
    \multicolumn{1}{|c|}{\multirow{4}{*}{NeWCRF~\cite{yuan2022neural}}} 
    & day   & 0.120  & 0.864 & 4.852 & 0.195 & 0.858 & 0.952 & 0.982 \\
          & night & 0.112 & 0.755 & 4.594 & 0.179 & 0.875 & 0.961 & 0.985 \\
          & rain  & 0.115 & 1.352 & 5.956 & 0.240  & 0.795 & 0.929 & 0.970 \\
          & \cellcolor{gray}avg&\cellcolor{gray}0.127 &\cellcolor{gray}0.965 &\cellcolor{gray}5.077 &\cellcolor{gray}0.202 &\cellcolor{gray}0.846 &\cellcolor{gray}0.949 &\cellcolor{gray}0.980 \\
    \hline
    \multicolumn{1}{|c|}{\multirow{4}{*}{DETI~\cite{shin2023deep} (mono)}} & day   & 0.115 & 0.983 & 4.895 & 0.201 & 0.882 & 0.952 & 0.977 \\
          & night & 0.107 & 0.850  & 4.658 & 0.185 & 0.894 & 0.961 & 0.981 \\
          & rain  & 0.152 & 1.567 & 6.020  & 0.247 & 0.822 & 0.928 & 0.964 \\
          &\cellcolor{gray}avg &\cellcolor{gray}0.123 &\cellcolor{gray}1.103 &\cellcolor{gray}5.134 &\cellcolor{gray}0.208 &\cellcolor{gray}0.869 &\cellcolor{gray}0.948 &\cellcolor{gray}0.975 \\
    \hline
    \multicolumn{1}{|c|}{\multirow{4}{*}{DETI~\cite{shin2023deep} (stereo)}} & day   & 0.113 & 0.948 & 4.852 & 0.200 & 0.884 & 0.953 & 0.977 \\
          & night & 0.105 & 0.811  & 4.584 & 0.183 & 0.896 & 0.961 & 0.981 \\
          & rain  & 0.149 & 1.499 & 5.940  & 0.245 & 0.826 & 0.929 & 0.965 \\
          & \cellcolor{gray}avg &\cellcolor{gray}0.120 &\cellcolor{gray}1.057 &\cellcolor{gray}5.068 &\cellcolor{gray}0.207 &\cellcolor{gray}\textbf{0.872} & \cellcolor{gray}0.949 &\cellcolor{gray}0.975 \\
    \hline
    \multicolumn{1}{|c|}{\multirow{4}{*}{Ours}} & day   & 0.098 & 0.549 & 3.593 & 0.139 & 0.893 & 0.980  & 0.996 \\
          & night & 0.103 & 0.519 & 3.398 & 0.142 & 0.888 & 0.980  & 0.995 \\
          & rain  & 0.130  & 0.802 & 4.461 & 0.173 & 0.830  & 0.968 & 0.993 \\
          &\cellcolor{gray}avg &\cellcolor{gray}\textbf{0.110} &\cellcolor{gray}\textbf{0.623} &\cellcolor{gray}\textbf{3.817} &\cellcolor{gray}\textbf{0.151} &\cellcolor{gray}0.870 &\cellcolor{gray}\textbf{0.976} &\cellcolor{gray}\textbf{0.995} \\
    \hline
    \end{tabular}%
}
  \label{tab:result}%
\end{table*}%

\begin{figure*}[ht]
    \centering
    \subfigure{
    \includegraphics[width=0.95\linewidth]{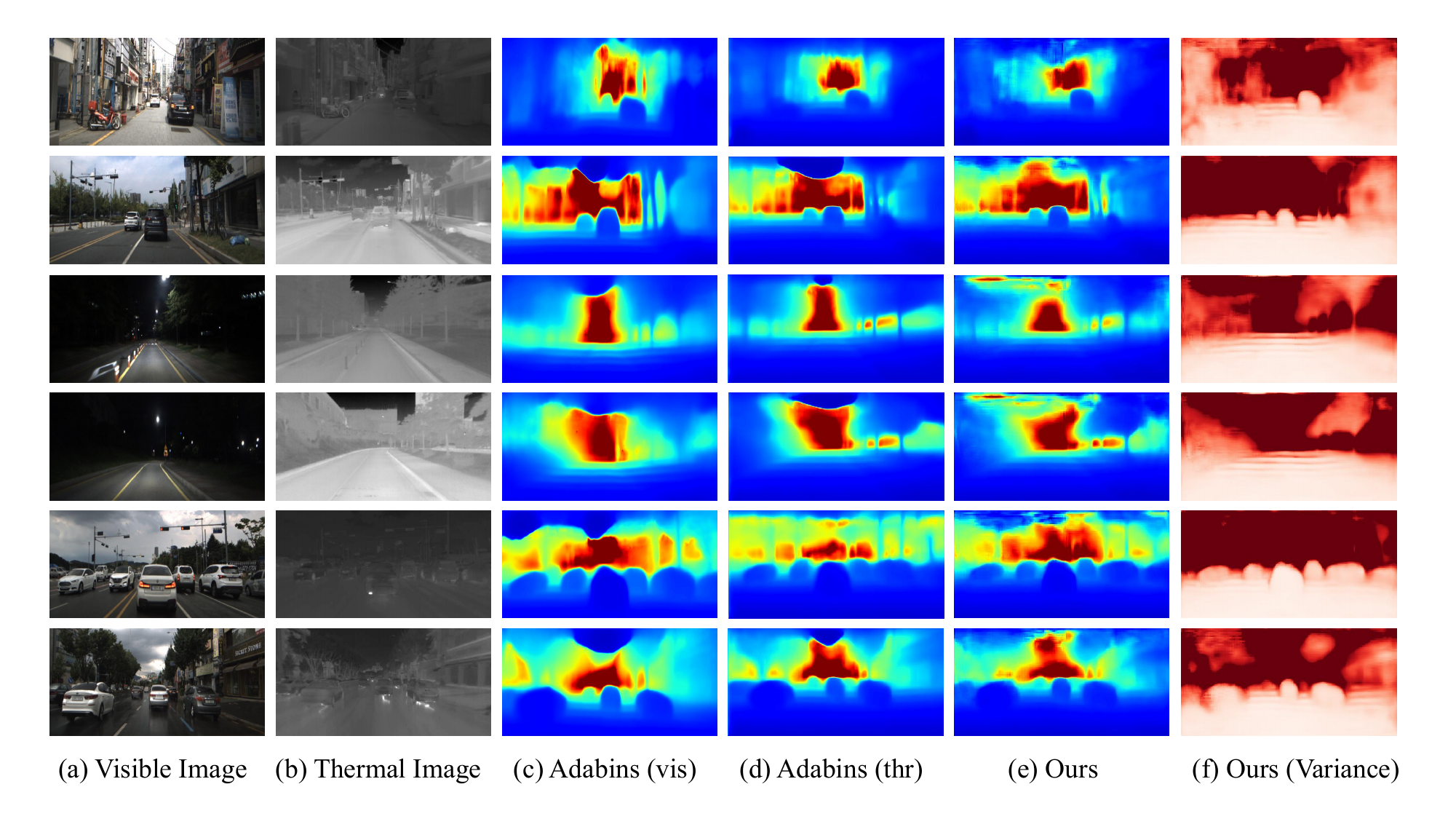}
    }
    
    \caption{\textbf{Quantitative depth comparison on the MS2 dataset.} From left to right: visible images, thermal images, depth maps generated by Adabins~\cite{bhat2021adabins} using either visible or thermal images, and depth and variance maps produced by our approach.The first two rows show results from the day test set, the middle two from the night test set, and the last two from the rainy test set. The results demonstrate that our method effectively leverages information from different modalities, producing robust and stable results under varying lighting conditions.}
    \label{fig:result}
    \vspace{-15pt}
\end{figure*}

\subsection{Datasets}

We utilize Multi-Spectral Stereo (MS2) benchmark dataset~\cite{shin2023deep} to evaluate our proposed method. MS2 dataset~\cite{shin2023deep} consists of approximately 195,000 pairs of multi-modal data, including stereo visible light images, stereo near-infrared (NIR) images, stereo long-wave infrared (thermal) images, stereo LiDAR point clouds, and GNSS/IMU information. We follow the official data split and conduct evaluations on test sets corresponding to different weather and lighting conditions, including clear daytime, nighttime, and rainy weather. 

\subsection{Implementation Details}
We implement our network with PyTorch~\cite{paszke2019pytorch} and conduct training on two NVIDIA RTX 4090 GPUs.  We use AdamW optimizer~\cite{loshchilov2017decoupled} and schedule the learning rate using 1cycle policy~\cite{smith2019super} with \(lr_{max}=3.57 \times 10^{-5}\) across all three modules. The batch size is 16/8/4 for MDP Module, CFM Module and Depth Module respectively. The number of epochs is 5 for all three modules. The input raw thermal image is transformed according to (\ref{eq:thermal}):
\begin{equation}
    \label{eq:thermal}
    T_{\text{Celsius}} = \frac{B}{\log\left(\frac{R}{\text{Raw} - O} + F\right)} - 273.15,
\end{equation}
where \( R = 380747 \), \( B = 1428 \), \( F = 1 \), and \( O = -88.539 \).

\subsection{Results}
\label{sec:benchmark}
Since there are very few methods for multi-spectrum stereo depth estimation, we compare our proposed method with the state-of-the art monocular and stereo depth networks DETI(mono) and DETI(stereo)~\cite{shin2023deep} from thermal images. 

The objective comparison results are shown in Table~\ref{tab:result}. We leverage the standard evaluation protocol from ~\cite{bhat2021adabins,bhat2022localbins,shin2023deep} to validate the efficacy of the proposed method in experiments, i.e., relative absolute error (Abs Rel), relative squared error (Sq Rel), root mean squared error (RMSE), root mean squared logarithmic error (RMSE log) and threshold accuracy (\(\delta < 1.25\), \(\delta < 1.25^2\), \(\delta < 1.25^3\)). Due to improvements in multi-modal information fusion and dual-view geometric constraints, our method achieves state-of-the-art performance on the MS2 dataset for most metrics, demonstrating a significant advantage in the Abs Rel metric. This is attributed to the successful integration of information across different modalities, which substantially reduces depth estimation errors. However, thermal images are significantly impacted by temperature variations, leading to degraded imaging quality in rainy conditions and resulting in some performance loss for our method.

The results of the subjective comparison are shown in Fig.~\ref{fig:result}.  Since DETI~\cite{shin2023deep} is not open-sourced, we present a comparison of subjective results using Adabins~\cite{bhat2021adabins}, trained on visible light or thermal images. It can be seen that the methods based on the visible light modality can provide accurate depth estimation when sufficient light is guaranteed. However, when lighting conditions deteriorate (e.g., at night or in rainy weather), their performance degrades rapidly. In contrast, methods based on thermal images, while maintaining relatively high visibility under adverse light conditions, face limitations due to their inherently lower resolution and contrast. This results in the generation of less detailed depth maps. Additionally, the dependence of thermal images on temperature means that their imaging quality can be further compromised in scenarios such as rainy weather, where temperature variations may affect the thermal signature. Our method integrates detailed information from visible light images with the enhanced visibility of thermal images in low-light scenarios, enabling it to generate accurate and robust depth maps across a wide range of lighting conditions.


\subsection{Ablation}

\textbf{Ablation study on Cross-Modal Feature Matching:} To investigate the importance of the CFM Module, we train two independent MDP Modules for visible light and thermal images, separately, and compare their performance with the full pipeline model. As shown in Table~\ref{tab:ablation_cmf}, it can be observed that our approach effectively leverages the advantages of both modalities, utilizing the geometric properties between the two views to further improve the accuracy of depth estimation. Additionally, as shown in Fig.~\ref{fig:fig1}, the ablation results demonstrate that without the CFM Module, the model's ability to capture fine-grained spatial details is notably diminished, highlighting the importance of cross-modal interaction for achieving superior depth perception.

\begin{table}[t!]

\setlength\tabcolsep{8.3pt} 
  \centering
  \caption{Ablation study on cross-modal feature matching}
    \begin{tabular}{|c|c|cc|c|}
    \hline
    \multirow{2}{*}{\textbf{Method}} & \multirow{2}{*}{\textbf{Condition}} & \multicolumn{2}{c|}{\textbf{Error} \(\downarrow\)} & \textbf{Accuracy} \(\uparrow\) \\
    \cline{3-5}
    & & Abs Rel & RMSE & \(\delta < 1.25\) \\
    \hline
    \multirow{4}{*}{Mono-VIS}
          & day   & 0.132 & 4.558 & 0.843 \\
          & night & 0.190  & 5.671 & 0.729 \\
          & rain  & 0.164 & 5.623 & 0.764 \\
          & \cellcolor{gray}avg  &\cellcolor{gray}0.162 &\cellcolor{gray}5.284 & \cellcolor{gray}0.779 \\ 
    \hline
    \multirow{4}{*}{Mono-THR}  
          & day   & 0.101 & 3.777 & 0.883 \\
          & night & 0.108 & 3.539 & 0.877 \\
          & rain  & 0.138 & 4.821 & 0.808 \\
          &\cellcolor{gray}avg&\cellcolor{gray}0.116 &\cellcolor{gray}4.046 &\cellcolor{gray}0.856 \\
    \hline
    \multirow{4}{*}{Stereo-MS}  
          & day   & 0.098 & 3.593 & 0.893 \\
          & night & 0.103 & 3.398 & 0.888 \\
          & rain  & 0.130  & 4.461 & 0.830 \\
          & \cellcolor{gray}avg &\cellcolor{gray}\textbf{0.110} &\cellcolor{gray}\textbf{3.817} & \cellcolor{gray}\textbf{0.870} \\
    \hline
    \end{tabular}%
  \label{tab:ablation_cmf}%
\end{table}%


\textbf{Ablation study on Degradation Masking:} We conduct an ablation study to evaluate the effect of Degradation Masking. Specifically, we retrain a network using only the CFM Module and Depth Module and compare its performance with our full model. The results can be found in Table~\ref{tab:addlabel}. Our model significantly outperforms the retrained network across all metrics, demonstrating the effectiveness of the proposed Degradation Masking.

\begin{table}[t!]
  \centering
  \caption{Ablation study on Degradation Mask}
    \begin{tabular}{|c|c|cc|c|}
    \hline
    \multirow{2}{*}{\textbf{Method}} & \multirow{2}{*}{\textbf{Condition}} & \multicolumn{2}{c|}{\textbf{Error} \(\downarrow\)} & \textbf{Accuracy} \(\uparrow\) \\
    \cline{3-5}
    & & Abs Rel & RMSE & \(\delta < 1.25\) \\
    \hline
    \multirow{4}{*}{Without Degration} & day   & 0.151 & 3.838 & 0.876 \\
          & night & 0.147 & 3.702 & 0.871 \\
          & rain  & 0.178 & 4.540  & 0.823 \\
          & \cellcolor{gray}avg &\cellcolor{gray}0.159 &\cellcolor{gray}4.027 &\cellcolor{gray}0.857 \\
    \hline
    \multirow{4}{*}{Full} & day   & 0.098 & 3.593 & 0.893 \\
          & night & 0.103 & 3.398 & 0.888 \\
          & rain  & 0.130  & 4.461 & 0.830 \\
          & \cellcolor{gray}avg & \cellcolor{gray}\textbf{0.110} &\cellcolor{gray}\textbf{3.817} & \cellcolor{gray}\textbf{0.870} \\
    \hline
    \end{tabular}%
  \label{tab:addlabel}%
\end{table}%

\vspace{-10pt}
\section{CONCLUSION}

In this paper, we propose a novel framework that integrates thermal and visible light images to produce accurate and robust depth maps across various lighting conditions. Specifically, we introduce Cross-Modal Feature Matching to bridge the gap between thermal and visible light images in depth estimation. Additionally, we present Degradation Masking to handle regions where matching fails due to insufficient lighting or texture loss in monocular thermal depth estimation. Our method achieves state-of-the-art performance on the MS2~\cite{shin2023deep} dataset. Ablation studies demonstrate that both the cross-modal matching mechanism and the degradation masking significantly enhance the precision and robustness of the algorithm.

\bibliographystyle{IEEEtran}
\bibliography{referencemc}
\end{document}